\documentclass[journal,onecolumn,12pt]{IEEEtran}
\date{}

\usepackage{amsfonts}
\usepackage{amsmath}
\usepackage{amssymb}
\usepackage{amsthm}
\usepackage{array}
\usepackage{caption}
\usepackage{cite}
\usepackage{citesort}
\usepackage{color}
\usepackage{epsfig}
\usepackage{enumitem} 
\usepackage{flexisym}
\usepackage{graphicx}
\usepackage{hyphenat}
\usepackage{ifpdf}
\usepackage{latexsym}
\usepackage{mathtools}
\usepackage{multirow}
\usepackage{psfrag}
\usepackage{setspace}
\usepackage{subfigure}
\usepackage{tikz}
\usepackage{times}
\usepackage{tkz-berge}
\usepackage{xparse}

\title{Scaled Nuclear Norm Minimization for Low-Rank Tensor Completion}

\author{Morteza Ashraphijuo and Xiaodong Wang \thanks{The authors are with the Department of Electrical Engineering, Columbia University, NY, email: \{ashraphijuo,wangx\}@ee.columbia.edu. }}
 
\begin{document}
\maketitle

\begin{abstract}

Minimizing the nuclear norm of a matrix has been shown to be very efficient in reconstructing a low-rank sampled matrix. Furthermore, minimizing the sum of nuclear norms of matricizations of a tensor has been shown to be very efficient in recovering a low-Tucker-rank sampled tensor. In this paper, we propose to recover a low-TT-rank sampled tensor by minimizing a weighted sum of nuclear norms of unfoldings of the tensor. We provide numerical results to show that our proposed method requires significantly less number of samples to recover to the original tensor in comparison with simply minimizing the sum of nuclear norms since the structure of the unfoldings in the TT tensor model is fundamentally different from that of matricizations in the Tucker tensor model.

\


\end{abstract}

\section{Introduction}

Tensors are generalizations of vectors and matrices to higher dimensions. Due to the recent advancement in machine learning, multi-dimensional analysis of data has become indispensable to fully exploit the high-dimensional representation of data as the conventional matrix analysis has only limited capability in exploiting correlations across different attributes in a multi-way representation. The low-rank tensor completion problem refers to completing a tensor given a subset of its entries and the corresponding rank constraints. There exists an extensive literature on the low-rank matrix completion problem, which is a special case of low-rank tensor completion problem (two-dimensional version). In general, there are many applications of low-rank tensor completion in various areas including image or signal processing \cite{phase,Image}, data mining \cite{data}, network coding \cite{network}, compressed sensing \cite{lim,sid,gandy}, reconstructing the visual data \cite{visual,liulow2,liu2016low}, seismic data processing \cite{kreimer,ely20135d,wang2016tensor}, etc. Tensors representing the real-world datasets usually exhibit a low rank structure and effectively exploiting such structure for analyzing large-scale high-dimensional datasets has become a hot topic in machine learning and data mining.

The majority of the literature on low-rank tensor completion is based on convex relaxation of matrix rank \cite{candes,candes2,cai,ashraphijuo2016c,phase} or different convex relaxations of tensor ranks \cite{gandy,tomioka,nuctensor,romera,hble,kreimer}. In addition, other approaches have been proposed that are based on alternating minimization \cite{wang2016tensor,liulow2,liu2016low}, algebraic geometric analyses \cite{charact,ashraphijuo4,ashraphijuo,ashraphijuo2,ashcon1,ashcon2,ashraphijuo3,ashraphijuo5} and other heuristics \cite{7347424,low,low2,goldfarb}. 
There are several well-known tensor decompositions including tensor-train (TT) decomposition \cite{oseledets,holtz}, Tucker decomposition \cite{Tuck,Tuckermanifold}, canonical polyadic (CP) decomposition \cite{harstions,kruskal}, tubal rank decomposition \cite{kilmer2013third}, etc. In this paper, we focus on TT-rank and TT decomposition. TT decomposition  was proposed in the field of quantum physics about $20$ years ago \cite{beck2n,scholy}. Later it was used in the area of machine learning \cite{oseledets,ose009king,oselesor}. A comprehensive survey on TT decomposition and the manifold of tensors of fixed TT rank can be found in \cite{TT} that also includes a comparison between the TT and Tucker decompositions for a better understanding of the advantages of TT decomposition.

The nuclear norm minimization for matrix completion problem, proposed in \cite{candes2}, can recover the original low-rank sampled matrix under some mild assumptions. The minimization of the sum of nuclear norms of matricizations of the tensor, proposed in \cite{visual}, can recover the original low-Tucker-rank sampled tensor under some mild assumptions \cite{hble}. One natural extension is to use the sum of nuclear norms of unfoldings to obtain the low-TT-rank sampled tensor. In this paper, we propose to use a weighted sum of nuclear norms of unfoldings, which outperforms the simple sum of nuclear norms of unfoldings. The reason behind such performance gain is the difference between the structure of matricizations in Tucker model and that of unfoldings in TT model. 

\section{Background on Low-TT-Rank Tensor Completion}\label{main}

Assume that a $d$-way tensor $\mathcal{U} \in \mathbb{R}^{n  \times \cdots \times n}$ is sampled. Denote $\Omega$ as the binary sampling pattern tensor that is of the same size as $\mathcal{U}$ and $\Omega(\vec{x})=1$ if $\mathcal{U}(\vec{x})$ is observed and  $\Omega(\vec{x})=0$ otherwise, where $\mathcal{U}(\vec{x})$ represents an entry of tensor $\mathcal{U}$  with coordinate $\vec{x}=(x_1,\dots,x_d)$.

Define the matrix $\mathbf{\widetilde U}_{(i)} \in \mathbb{R}^{n^i \times  n^{d-i}}$ as the $i$-th {\it unfolding} of the tensor $\mathcal{U}$, such that $\mathcal{U}(\vec{x}) = \\ \mathbf{\widetilde U}_{(i)}({\widetilde  M}_{i} (x_1,\dots,x_i),{\widetilde{{M}} }_{-i} (x_{i+1},\ldots,x_d))$, where ${\widetilde M}_{i}: (x_1,\dots,x_i) \rightarrow  \{1,2,\dots, n^i\}$ and ${\widetilde{{M}}}_{-i}: (x_{i+1},\ldots,x_d)  \rightarrow  \{1,2,\dots, n^{d-i} \}$ are two bijective mappings.

The separation or TT-rank of a tensor is defined as $\text{rank}_{\text{TT}} (\mathcal{U})=(u_1,\ldots,u_{d-1})$ where $u_i = \text{rank}(\mathbf{\widetilde U}_{(i)})$, $i=1,\dots,d-1$. Note that $u_i \leq \min \{n^i, n^{d-i} \}$ in general and also $r_1$ is simply the conventional matrix rank when $d=2$. The TT decomposition of a tensor $\mathcal{U}$ is defined as
\begin{eqnarray}\label{TTeq2}
\mathcal{U}(\vec{x}) =  \sum_{k_1=1}^{u_1} \cdots \sum_{k_{d-1}=1}^{u_{d-1}}  \mathcal{U}^{(1)}(x_1,k_1)  \left( \prod_{i=2}^{d-1} \mathcal{U}^{(i)}(k_{i-1},x_i,k_i) \right) \mathcal{U}^{(d)}(k_{d-1},x_d).
\end{eqnarray}
or in short,
\begin{eqnarray}\label{TTeq1}
\mathcal{U} = \mathcal{U}^{(1)} \dots  \mathcal{U}^{(d)},
\end{eqnarray}
where the $3$-way tensors $\mathcal{U}^{(i)} \in \mathbb{R}^{u_{i-1} \times n_i \times u_{i}}$ for $i=2,\dots,d-1$ and matrices $\mathcal{U}^{(1)} \in \mathbb{R}^{n_1 \times u_1}$  and $\mathcal{U}^{(d)} \in \mathbb{R}^{u_{d-1} \times n_d}$ are the components of this decomposition.

Let $\mathbf{U}_{(i)}$ be the $i$-th {\it matricization} of the tensor $\mathcal{U}$, i.e., the matrix $\mathbf{U}_{(i)} \in \mathbb{R}^{n \times n^{d-1}}$ such that $\mathcal{U}(\vec{x}) = {\mathbf{U}}_{(i)}(x_i,{M}_{i} (x_1,\ldots,x_{i-1},x_{i+1},\ldots,x_d))$, where ${M}_{i}: (x_1,\ldots,x_{i-1},x_{i+1},\ldots,x_d) \rightarrow \{1,2,\dots, n^{d-1} \}$  is a bijective mapping. Observe that for any arbitrary tensor $\mathcal{A}$, the first matricization and the first unfolding are the same, i.e., $\mathbf{ A}_{(1)} = \mathbf{\widetilde A}_{(1)}$. The Tucker-rank  of a tensor is defined as $\text{rank}_{\text{Tucker}} (\mathcal{U})=(m_1,\ldots,m_d)$ where $m_i = \text{rank}(\mathbf{U}_{(i)})$.

Define $\mathcal{U}_{{\Omega}}$ as the tensor obtained from sampling $\mathcal{U}$ according to ${\Omega}$, i.e.,
\begin{eqnarray}
\mathcal{U}_{{\Omega}} (\vec{x}) = \left\{
	\begin{array}{ll}
		\mathcal{U} (\vec{x})  & \mbox{if } \ {\Omega}(\vec{x}) \ = 1, \\
		0  & \mbox{if } \ {\Omega}(\vec{x}) \ = 0.
	\end{array}
\right.
\end{eqnarray}

Assuming that a tensor $\mathcal{U}$ with $\text{rank}_{\text{TT}} (\mathcal{U})=(u_1,\dots,u_{d-1})$ is sampled according to the sampling pattern $\Omega$. Then, the following NP-hard problem, known as the rank feasibility problem, or tensor completion problem, aims to find a completion of the given rank constarints.

\begin{align}\label{ouiho}
& \ \ \ \ \ \  \ \ \ \ \ \  \   \ \text{find}_{\mathcal{U}^{\prime} \in \mathbb{R}^{n \times \cdots \times n}} 
& &  \mathcal{U}^{\prime} \ \ \ \ \  \ \ \  \ \ \  \ \ \  \ \ \  \ \ \  \ \ \  \ \ \  \ \ \   \\
& \ \ \ \ \ \  \ \ \ \ \ \  \   \ \text{subject to}
& & \mathcal{U}^{\prime}_{{\Omega}} = \mathcal{U}_{{\Omega}}, \nonumber \\
& & & \text{rank}_{\text{TT}} (\mathcal{U}^{\prime}) = \text{rank}_{\text{TT}} (\mathcal{U}). \nonumber
\end{align}

\section{Optimization Formulations}\label{optmai}

As mentioned earlier, for the matrix case, by relaxing the rank constraint and minimizing the nuclear norm of the matrix, we can obtain the original low-rank matrix under some mild assumptions \cite{candes2}. Following this idea, the same problem for low-Tucker-rank tenors is studied in \cite{visual}, where by relaxing the Tucker-rank constraints and minimizing the sum of nuclear norms of matricizations of the tensor, the low-Tucker-rank sampled tensor can be obtained \cite{hble}. This formulation can be written as
\begin{align}\label{ombmn}
& \ \ \ \ \ \  \ \ \ \ \ \  \   \ \text{minimize}_{\mathcal{U}^{\prime} \in \mathbb{R}^{n \times \cdots \times n}} 
& &  \sum_{i=1}^{d} \|\mathbf{ U^{\prime}}_{(i)}\|_{*} \ \ \ \ \  \ \ \  \ \ \  \ \ \  \ \ \  \ \ \  \ \ \  \ \ \  \ \ \   \\
& \ \ \ \ \ \  \ \ \ \ \ \  \   \ \text{subject to}
& & \mathcal{U}^{\prime}_{{\Omega}} = \mathcal{U}_{{\Omega}}, \nonumber
\end{align}
where $\|\mathbf{X}\|_{*}$ denotes the nuclear norm of the matrix $\mathbf{X}$, i.e., sum of the singular values of $\mathbf{X}$. Intuitively, this optimization formulation is not efficient for solving \eqref{ouiho} as the tensor is chosen generically from the corresponding low-TT-rank manifold and is not a low-Tucker-rank tensor with high probability. In the numerical experiments, we show that this method performs very poorly. Now, given that the TT-rank is defined through the unfoldings, a natural alternative formulation is to minimize the sum of nuclear norms of unfoldings as
\begin{align}\label{omjasn}
& \ \ \ \ \ \  \ \ \ \ \ \  \   \ \text{minimize}_{\mathcal{U}^{\prime} \in \mathbb{R}^{n \times \cdots \times n}} 
& &  \sum_{i=1}^{d-1} \|\mathbf{\widetilde U^{\prime}}_{(i)}\|_{*} \ \ \ \ \  \ \ \  \ \ \  \ \ \  \ \ \  \ \ \  \ \ \  \ \ \  \ \ \   \\
& \ \ \ \ \ \  \ \ \ \ \ \  \   \ \text{subject to}
& & \mathcal{U}^{\prime}_{{\Omega}} = \mathcal{U}_{{\Omega}}. \nonumber
\end{align}

In the numerical experiments, we show that this method performs much better than the previous formulation \eqref{ombmn}, which is very reasonable as we are minimizing the tightest convex relaxation of each TT-rank component. On the other hand, since the dimensions of different unfoldings are different, i.e., $\mathbf{\widetilde U^{\prime}}_{(i)} \in \mathbb{R}^{n^i \times n^{d-i}}$, an even more efficient formulation is to use the following weighted sum of nuclear norms 
\begin{align}\label{khfeb}
& \ \ \ \ \ \  \ \ \ \ \ \  \   \ \text{minimize}_{\mathcal{U}^{\prime} \in \mathbb{R}^{n \times \cdots \times n}} 
& &  \sum_{i=1}^{d-1} \min \{n^i, n^{d-i} \} \|\mathbf{\widetilde U^{\prime}}_{(i)}\|_{*} \ \ \ \ \  \ \ \  \ \ \  \ \ \  \ \ \  \ \ \  \ \ \  \ \ \  \ \ \   \\
& \ \ \ \ \ \  \ \ \ \ \ \  \   \ \text{subject to}
& & \mathcal{U}^{\prime}_{{\Omega}} = \mathcal{U}_{{\Omega}}. \nonumber
\end{align}

Note that all three optimization formulations \eqref{ombmn}, \eqref{omjasn} and \eqref{khfeb} are convex programs and easy to solve.


\section{Numerical Results}

In our numerical experiments, we first generate a generic tensor $\mathcal{U}$ of a given TT rank as the following. We consider a TT-rank vector $(u_1,\dots,u_{d-1})$ and we generate completely random two and three-way tensor components $\mathcal{U}^{(1)} \in \mathbb{R}^{n_1 \times u_1}, \mathcal{U}^{(2)} \in \mathbb{R}^{u_1 \times n_2 \times u_2}, \dots,  \mathcal{U}^{(d)} \in \mathbb{R}^{u_{d-1} \times n_d}$ and construct $\mathcal{U}$ according to \eqref{TTeq2}. Hence, $\mathcal{U}$ is generically chosen from the manifold of tensors of TT-rank $(u_1,\dots,u_{d-1})$. Moreover, we sample the entries of the obtained tensor $\mathcal{U}$ independently and with some probability $p$.

\begin{figure}[h]
	\centering
		{\includegraphics[width=11.4cm]{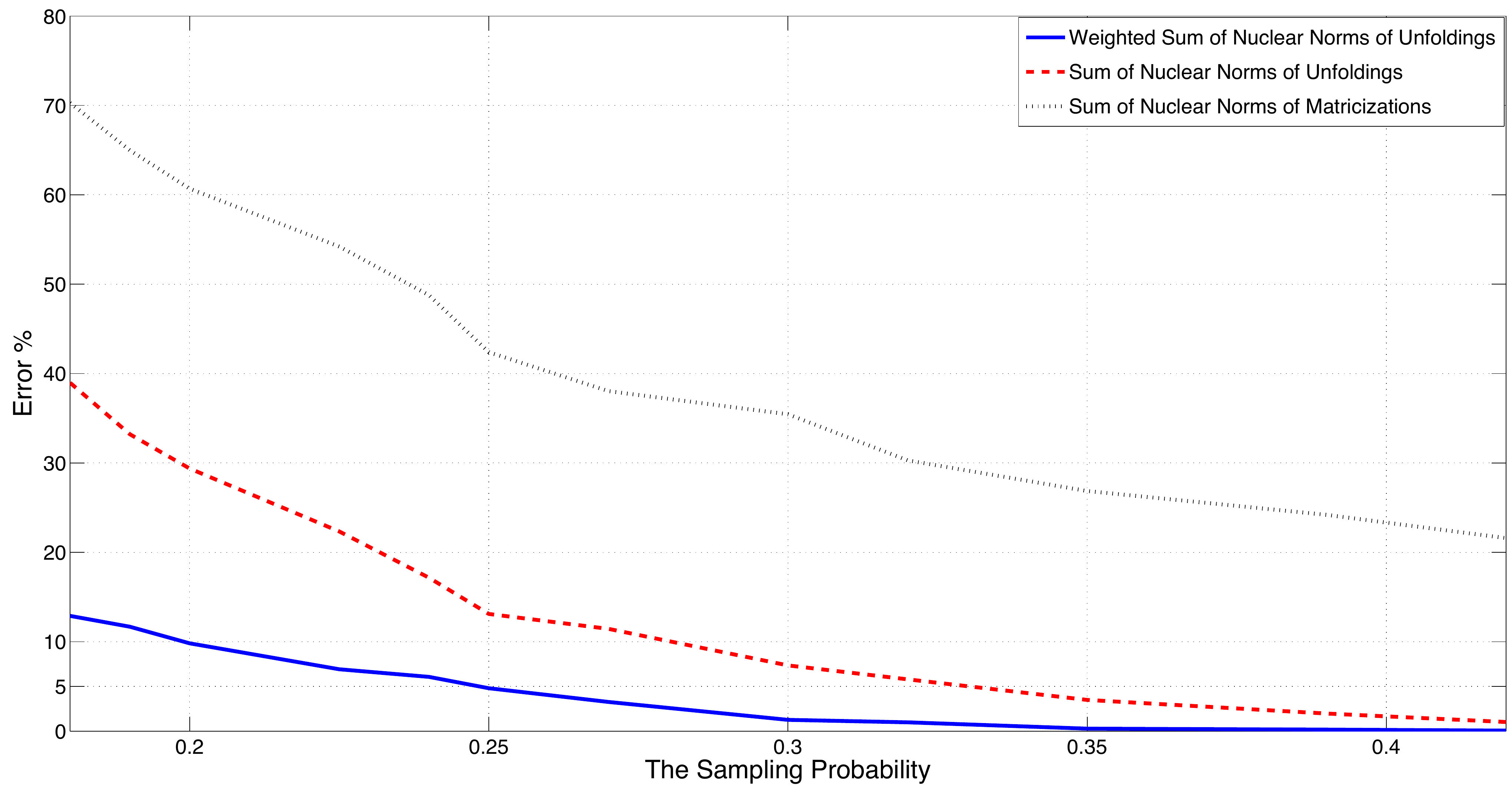}}
	\caption{ Comparison of formulations \eqref{ombmn}, \eqref{omjasn} and \eqref{khfeb} for $\mathcal{U} \in \mathbb{R}^{4 \times 4 \times 4 \times 4 \times 4}$ of TT-rank $(1,4,2,2)$.}
	\label{fig1}\vspace{-4mm}
\end{figure}

For the first example, we construct a generic tensor $\mathcal{U} \in \mathbb{R}^{4 \times 4 \times 4 \times 4 \times 4}$ of TT-rank $(1,4,2,2)$ and sample each entry with probability $p$. Then, we solve each one of the optimization problems \eqref{ombmn}-\eqref{khfeb} for the sampled tensor to reconstruct the original tensor. We define the error as $\frac{\|\mathcal{\hat U} - \mathcal{U}\|}{\|\mathcal{U}\|} \times 100\%$, where $\mathcal{\hat U}$ is the obtained solution and $\mathcal{U}$ is the original sampled tensor. In Figure \ref{fig1}, we plot the errors obtained from \eqref{ombmn}, \eqref{omjasn} and \eqref{khfeb} in terms of the sampling probability. For this experiment, we repeated each experiment $100$ times for each value of the sampling probability $p$ and the error curves represent the average over the $100$ experiments.


For example, according to Figure \ref{fig1}, using our proposed weighted sum of nuclear norms of unfoldings, the error of $1\%$ can be obtained for sampling probability $p=0.32$, whereas $p=0.42$ is needed for the same error using the sum of nuclear norms of unfoldings. In other words, our proposed method outperforms the method using the sum of nuclear norms of unfoldings by approximately $ 23.8 \%$ in terms of the sampling probability. Moreover, by decreasing the sampling probability, the sum of nuclear norms results in a much greater error in comparison with our proposed objective function. Note that formulation based on the sum of nuclear norms of matricizations performs very poorly.

\begin{figure}[h]
	\centering
		{\includegraphics[width=11.4cm]{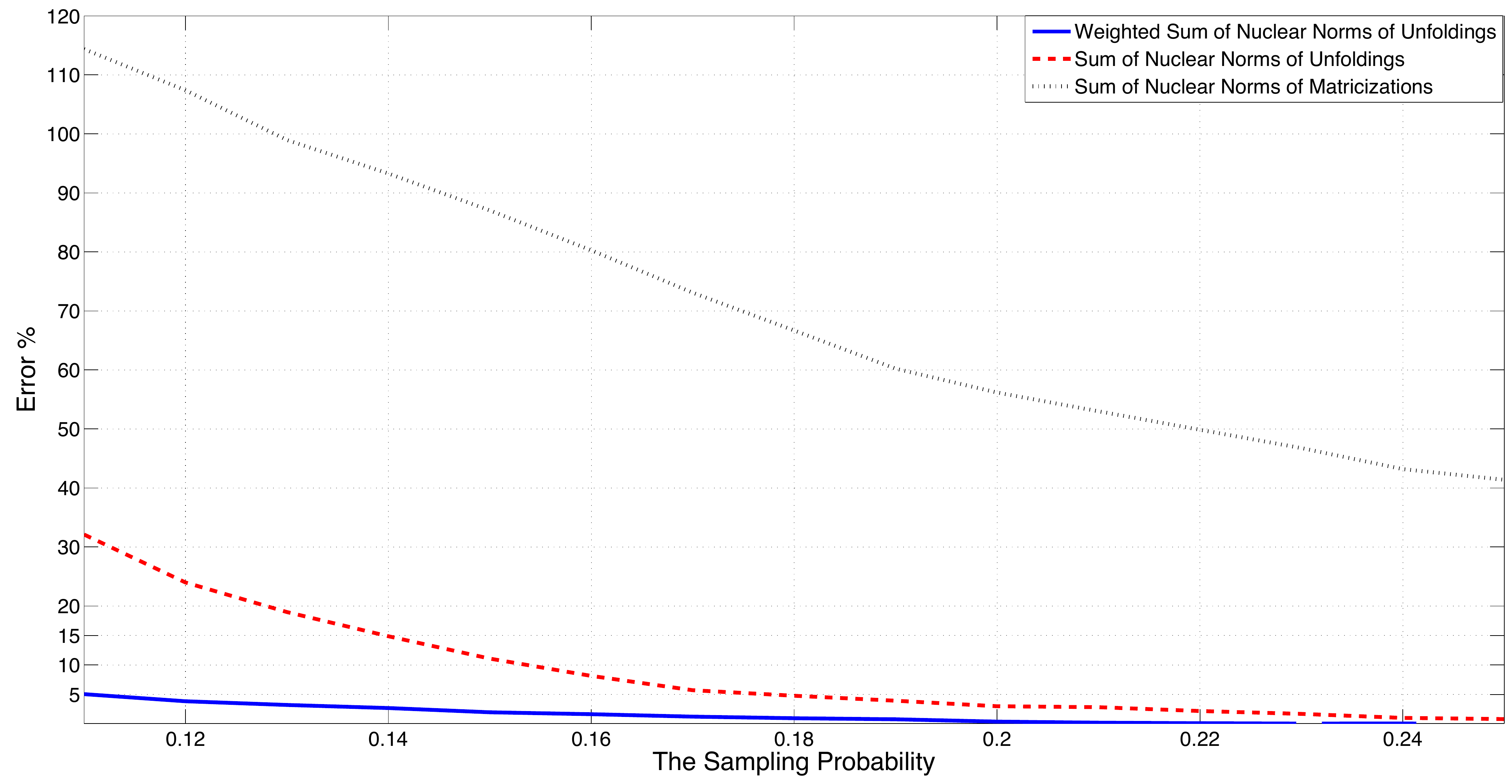}}
	\caption{ Comparison of formulations \eqref{ombmn}, \eqref{omjasn} and \eqref{khfeb} for $\mathcal{U} \in \mathbb{R}^{5 \times 5 \times 5 \times 5 \times 5}$ of TT-rank $(3,9,10,2)$.}
	\label{fig2}\vspace{-4mm}
\end{figure}

As the second example, in Figure \ref{fig2}, we represent the error obtained from \eqref{ombmn}, \eqref{omjasn} and \eqref{khfeb} for a sampled tensor $\mathcal{U} \in \mathbb{R}^{5 \times 5 \times 5 \times 5 \times 5}$ of TT-rank $(3,9,10,2)$. Using our proposed weighted sum of nuclear norms of unfoldings, the error of $1\%$ can be obtained for sampling probability $p=0.18$, whereas $p=0.25$ is needed for the same error using the sum of nuclear norms of unfoldings. Hence, our proposed method outperforms the method using the sum of nuclear norms of unfoldings by approximately $ 28 \%$ in terms of the sampling probability. Again, the sum of nuclear norms of matricizations performs very poorly. For this experiment, we repeated each experiment $100$ times for each value of the sampling probability $p$ and the error curves represent the average over the $100$ experiments.

\begin{figure}[h]
	\centering
		{\includegraphics[width=11.4cm]{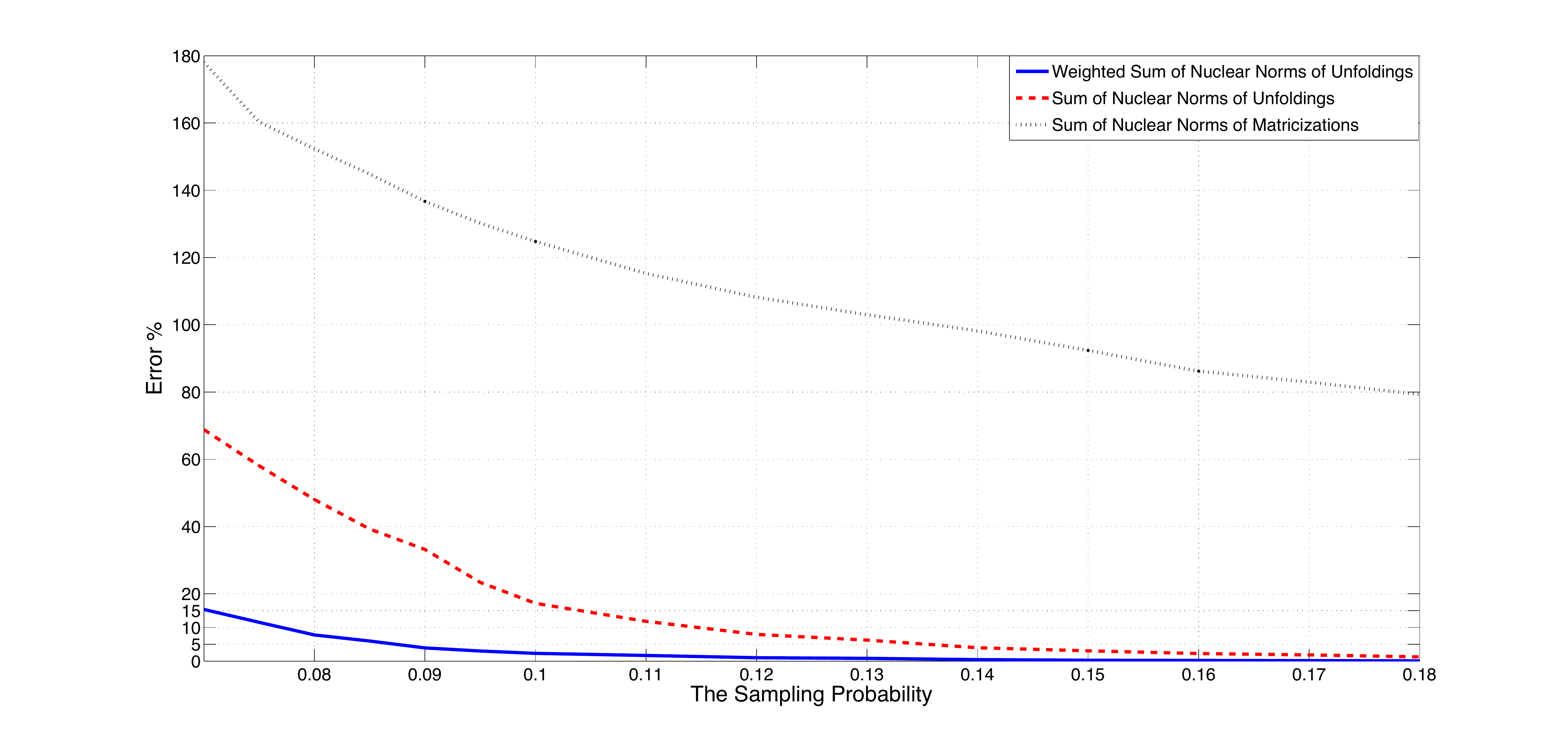}}
	\caption{ Comparison of formulations \eqref{ombmn}, \eqref{omjasn} and \eqref{khfeb} for $\mathcal{U} \in \mathbb{R}^{4 \times 4 \times 4 \times 4 \times 4 \times 4}$ of TT-rank $(1,5,15,10,3)$.}
	\label{fig3}\vspace{-4mm}
\end{figure}

Finally, in Figure \ref{fig3}, we represent the error obtained from \eqref{ombmn}, \eqref{omjasn} and \eqref{khfeb} for a sampled tensor $\mathcal{U} \in \mathbb{R}^{4 \times 4 \times 4 \times 4 \times 4 \times 4}$ of TT-rank $(1,5,15,10,3)$. Using our proposed weighted sum of nuclear norms of unfoldings, the error of $1\%$ can be obtained for sampling probability $p=0.12$, whereas $p=0.18$ is needed for the same error using the sum of nuclear norms of unfoldings. Hence, our proposed method outperforms the method using the sum of nuclear norms of unfoldings by approximately $33.3 \%$ in terms of the sampling probability. For this experiment, we repeated each experiment $50$ times for each value of the sampling probability $p$ and the error curves represent the average over the $50$ experiments.

\section{Conclusions}\label{conclusection}

Minimizing the nuclear norm of a matrix is a well-known and efficient method to tackle the low-rank matrix completion problem. However, the nuclear norm of a tensor is not well defined, and therefore one way to approach the low-rank tensor completion problem is to minimize the sum of nuclear norms of matricizations or unfoldings of the tensor. In fact, minimizing the sum of nuclear norms of matricizations of a tensor is efficient to recover a low-Tucker-rank sampled tensor. In order to recover a low-TT-rank sampled tensor, we proposed to minimize a weighted sum of nuclear norms of unfoldings of the tensor instead of minimizing the sum of nuclear norms of unfoldings. Through numerical results, we showed that our proposed optimization formulation outperforms the formulations using the sum of nuclear norms of unfoldings or matricizations significantly in the sense of the required number of samples to recover the original tensor.

\bibliographystyle{IEEETran}
\bibliography{bib}

\end{document}